\documentclass{styles/svproc}
\usepackage{url}

\usepackage{comment}
\usepackage{graphicx}
\usepackage{tikz}
\usetikzlibrary{trees}

\begin{document}
\mainmatter              
\title{
Design of a Human-Assistance Robot System with Contextual Action Recognition
}
\titlerunning{\textit{Proactive, Context-Aware, and Dynamic Robots for Human Environments}}  
%
\author{Amanuel Ergogo\inst{1,2} \and Teresa Zielińska\inst{1}}
\authorrunning{A.Ergogo et al.} 
%
%
\institute{Warsaw University of Technology, Warsaw, Poland\\
\and
SANO Centre for Computational Personalized Medicine, Krakow,
Poland \\
\email{a.ergogo@sanoscience.org, amanuel.ergogo.dokt@pw.edu.pl}
}

\maketitle              

\begin{abstract}
This paper presents a conceptual design for a proactive human-assisting robot system capable of recognizing human activities and responding proactively. The system leverages contextual human activity recognition to interpret human actions across diverse contexts, while behavior trees are utilized to define dynamic and interpretable robot behaviors. We outline the system architecture, incorporating contextual human action recognition (HAR), behavior trees (BTs), and ROS, using the Spot robot platform as a representative example. We explain how HAR enables the robot to provide proactive assistance, discuss its limitations, and introduce methodologies for contextual HAR to address these limitations, thereby enhancing the robot’s decision-making in complex human activity scenarios.
, behavior trees
\end{abstract}

\section{Introduction}

Advancements in computation and hardware have significantly propelled the field of human-centered robotics, leading to the development of assistive technologies that enhance human capabilities and quality of life~\cite{mataric2007socially}. Key research areas include human-robot interaction (HRI)~\cite{goodrich2008human}, human motion analysis~\cite{poppe2007vision}, and collaborative robotics~\cite{kruger2009cooperation}, all aimed at enabling robots to operate effectively in human environments.

Human-assisting robots are designed to work alongside humans, providing support in domains such as healthcare~\cite{brooks2010robots}, domestic assistance~\cite{srinivasa2010herb}, and industrial collaboration~\cite{villani2018survey}. For instance, assistive robots can help the elderly and disabled with daily tasks~\cite{tapus2007socially}, perform household chores~\cite{pratt2013darpa}, or collaborate with humans in manufacturing settings~\cite{huber2008human}. However, deploying robots in human spaces presents technical challenges due to the dynamic and unpredictable nature of these environments~\cite{thrun2004toward}.

A primary challenge is establishing effective HRI that allows robots to intuitively interpret and respond to human actions~\cite{fong2003survey}. Additionally, defining dynamic robot behaviors that can adapt to evolving environments is crucial~\cite{colledanchise2018behavior}. Traditional HRI approaches utilize voice commands~\cite{mutlu2012conversational}, gestures~\cite{nickel2007visual}, and input devices like keypads~\cite{beer2014toward}. For robot behavior definition, finite state machines are commonly employed~\cite{perez2011robot}. While these methods facilitate basic interaction and control, they often lack the flexibility and proactivity required for seamless integration into human environments.

Human Activity Recognition (HAR) focuses on automatically detecting and classifying human actions from sensory data~\cite{aggarwal2011human}. In robotics, HAR applications have been limited~\cite{chen2013survey}. Recent efforts employ machine learning models trained on data from wearables~\cite{lara2013survey} and cameras~\cite{weinland2011survey} to recognize human activities by mapping spatial and temporal features to activity classes. These models are effective for recognizing simple gestures but struggle with complex activities involving intricate human-object interactions and contextual nuances~\cite{kong2018human}.

To improve accuracy in activity recognition, HAR models need to incorporate contextual cues from the environment, understand human-object interactions, and consider user routines or preferences~\cite{rahmani2018learning}. Contextual HAR enhances a robot's ability to interpret human actions within situational contexts, leading to more appropriate and proactive responses.

In this paper, we propose a framework that integrates contextual HAR with BTs to dynamically define robot behaviors, enabling proactive assistance in complex human activity scenarios. Our contributions are:

\begin{enumerate}
    \item We outline a system architecture that incorporates contextual HAR, BTs, and ROS, using the Spot robot platform as an example.
    \item We demonstrate how contextual HAR enables the robot to provide proactive assistance, discuss its limitations, and introduce methodologies to address these challenges, enhancing decision-making in complex scenarios.
\end{enumerate}

By leveraging contextual HAR and BTs, our system aims to overcome the limitations of traditional HRI methods, providing a dynamic and flexible framework for human-assisting robots operating in dynamic environments.

\section{Related Work}
\subsection{Human action recognition}
\textit{Human activity recognition (HAR)} is essential for enabling robots to understand and interact with humans in dynamic environments. Traditional HAR methods relied on handcrafted features extracted from video data, such as silhouettes, optical flow, and spatiotemporal interest points \cite{poppe2010survey}. While these approaches achieved moderate success, they struggled with variations in actions and environmental conditions.

The advent of deep learning significantly advanced HAR by allowing models to learn features directly from data. Convolutional Neural Networks (CNNs) and Recurrent Neural Networks (RNNs) have been employed to capture spatial and temporal dynamics in videos \cite{ji20133d}. Ji et al. \cite{ji20133d} proposed 3D CNNs for action recognition, effectively capturing spatio-temporal features. Donahue et al. \cite{donahue2015long} introduced Long-term Recurrent Convolutional Networks (LRCN), combining CNNs and RNNs for end-to-end learning of video data.

Two-stream networks, processing RGB frames and optical flow separately, have been successful in capturing both appearance and motion information \cite{simonyan2014two}. However, these models often require substantial computational resources, limiting their applicability in real-time robotic systems.

Skeleton-based HAR has gained attention due to its efficiency and robustness to environmental variations. Methods using depth sensors and pose estimation represent human actions as sequences of skeletal joint positions \cite{du2015hierarchical}. Du et al. \cite{du2015hierarchical} proposed a hierarchical RNN for skeleton-based action recognition, modeling the human body structure. Yan et al. \cite{yan2018spatial} introduced Spatial Temporal Graph Convolutional Networks (ST-GCN), effectively capturing spatial and temporal dependencies in skeletal data.

Incorporating contextual information enhances the accuracy of HAR by considering environmental cues, object interactions, and scene semantics \cite{wang2016hierarchical}. Wang et al. \cite{wang2016hierarchical} developed a hierarchical attention network that focuses on relevant spatial and temporal regions based on context, improving action recognition performance.

In robotics, integrating HAR with contextual understanding enables robots to interpret human intentions and respond appropriately \cite{koppula2013anticipating}. Koppula and Saxena \cite{koppula2013anticipating} presented an approach for robots to anticipate human activities by modeling human-object interactions and affordances, facilitating proactive assistance.

Despite these advancements, challenges remain in deploying HAR in robotic systems. Many deep learning models are computationally intensive, posing difficulties for real-time processing on robots with limited resources. Additionally, models trained on benchmark datasets may not generalize well to unstructured environments encountered in real-world scenarios \cite{torralba2011unbiased}.

Our work addresses these challenges by leveraging contextual HAR to improve action recognition accuracy in dynamic environments and integrating it with Behavior Trees (BTs) for flexible robot behavior definition, enabling proactive human-assisting robotic systems.

\subsection{Designing robot behavior }

Designing robot behavior is crucial for enabling autonomous systems to operate effectively in dynamic environments. Traditional approaches often use Finite State Machines (FSMs) due to their simplicity and clear structure~\cite{gonzalez1998finite}. FSMs model behaviors as a set of states and transitions, suitable for straightforward and well-defined processes. However, FSMs face limitations when dealing with complex and dynamic tasks, as they can become unwieldy and difficult to manage with increasing complexity~\cite{colledanchise2018behavior}.

BBTs have emerged as a flexible alternative for modeling robot behaviors, offering advantages in modularity, scalability, and reusability~\cite{colledanchise2018behavior}. Originating from the gaming industry, BTs represent behaviors in a hierarchical tree structure comprising control flow nodes and execution nodes, allowing for more manageable and adaptable behavior definitions~\cite{isla2005handling}.

Compared to FSMs, BTs provide several benefits. Firstly, BTs enable the reuse of behavior subtrees across different tasks, promoting modular design and simplifying complex behaviors~\cite{colledanchise2017behavior}. Secondly, the hierarchical structure of BTs manages complexity more effectively than the flat structure of FSMs, reducing the likelihood of state explosion~\cite{colledanchise2018behavior}. Thirdly, BTs allow for dynamic modification of behaviors at runtime, which is challenging with traditional FSMs~\cite{marzinotto2014towards}. Key differences between BTs and FSMs are summarized in Table~\ref{tab:bt_vs_fsm}.

\begin{table}[h]
\centering
\caption{Comparison between Behavior Trees and Finite State Machines}
\label{tab:bt_vs_fsm}
\begin{tabular}{|p{1.9cm}|p{5cm}|p{5cm}|}
\hline
\textbf{Aspect} & \textbf{BTs} & \textbf{FSMs} \\ \hline
Structure & Hierarchical tree of nodes & Flat collection of states and transitions \\ \hline
Modularity & High; subtrees can be reused~\cite{colledanchise2017behavior} & Low; difficult to reuse states \\ \hline
Scalability & Scales well with complexity & Poor scalability; prone to state explosion~\cite{colledanchise2018behavior} \\ \hline
Flexibility & Allows dynamic modification at runtime~\cite{marzinotto2014towards} & Static; changes require redesign \\ \hline
Debugging & Easier due to hierarchical nature & Can be difficult with complex transitions \\ \hline
\end{tabular}
\end{table}

Several tools and libraries facilitate the implementation of BTs in robotics, each offering distinct features. BehaviorTree.CPP is an open-source C++ library designed for robotics applications~\cite{faconti2020behaviortreecpp}. It supports efficient execution, seamless integration with the Robot Operating System (ROS), and provides Groot, a graphical interface for designing and debugging BTs.

PyTrees is a Python library for implementing BTs~\cite{stonier2015pytrees}. It is user-friendly and ideal for rapid prototyping, with features including ease of use, ROS support via the \texttt{py\_trees\_ros} package, and visualization tools. Table~\ref{tab:bt_libraries} compares some of the prominent BT libraries.

\begin{table}[h]
\centering
\caption{Comparison of Behavior Tree Libraries}
\label{tab:bt_libraries}
\begin{tabular}{|p{3.3cm}|p{1.6cm}|p{2cm}|p{4cm}|}
\hline
\textbf{Library} & \textbf{Language} & \textbf{ROS Integration} & \textbf{Key Features} \\ \hline
BehaviorTree.CPP~\cite{faconti2020behaviortreecpp} & C++ & Yes & High performance, Groot visualization, asynchronous actions \\ \hline
PyTrees~\cite{stonier2015pytrees} & Python & Yes & Ease of use, rapid prototyping, ROS support \\ \hline
BehaviorTree.NET & C\# & Limited & Suitable for .NET applications \\ \hline
\end{tabular}
\end{table}

For our application, which requires efficient, real-time performance and ROS integration, BehaviorTree.CPP is the preferred choice. Its advantages include high performance due to its C++ implementation, support for asynchronous actions and multithreading, and advanced visualization and debugging tools. While PyTrees is suitable for prototyping and educational purposes, it may not meet the performance demands of our system.

Therefore, BehaviorTree.CPP provides the necessary features and efficiency required for dynamic robot behavior definition in complex environments.

\subsection{
            Selecting tools for spot programming
            }

The Boston Dynamics Spot robot is a quadruped robotic platform designed for versatile applications in inspection, data collection, and manipulation tasks~\cite{bostondynamics2020spot}. Spot is equipped with advanced mobility capabilities, allowing it to navigate complex terrains and environments. Key features include dynamic walking, stair-climbing, obstacle avoidance, modular payloads like cameras and sensors, and a robotic arm for manipulation tasks. The robot also possesses integrated cameras and sensors for environment mapping and object detection.

To access this spot robot capability there are various middlewares can be used, commonly spot software development kit ( spot SDK) and Spot ROS.
The Spot SDK enables developers to create custom applications for the robot~\cite{bostondynamics2020spotSDK}. The SDK utilizes gRPC, a high-performance, open-source Remote Procedure Call (RPC) framework developed by Google~\cite{grpc2020}. gRPC facilitates efficient communication between client applications and the robot, supporting multiple programming languages. It allows developers to send control commands for locomotion and manipulation, access sensor data and camera feeds in real time, and utilize built-in autonomous behaviors like obstacle avoidance and navigation.

The Robot Operating System (ROS) is an open-source framework widely used in robotics for its tools and libraries that facilitate the development of robot applications~\cite{quigley2009ros}. ROS provides a standardized communication infrastructure, simplifying the integration of various software components within a robotic system.

While the Spot SDK uses gRPC for communication, there is also a ROS driver available for Spot that enables integration with the ROS ecosystem. Using the Spot SDK with gRPC provides direct and efficient access to the robot's capabilities, suitable for applications requiring high performance and fine-grained control. The SDK supports multiple programming languages, offering flexibility in development.

Integrating Spot with ROS allows developers to leverage the extensive ecosystem of ROS tools and packages. This integration facilitates interoperability with other ROS-compatible robots and sensors and enables the use of standard ROS functionalities such as visualization with \texttt{rviz}, data recording with \texttt{rosbag}, and coordinate transformations with \texttt{tf}~\cite{quigley2009ros}. Table~\ref{tab:grpc_vs_ros} compares the two approaches.

\begin{table}[h]
\centering
\caption{Comparison between Spot SDK (gRPC) and ROS Integration}
\label{tab:grpc_vs_ros}
\begin{tabular}{|p{2.3cm}|p{4.5cm}|p{4.5cm}|}
\hline
\textbf{Aspect} & \textbf{Spot SDK} & \textbf{Spot ROS integration} \\ \hline
Communication Protocol & gRPC-based API & ROS messages and services \\ \hline
Language Support & Multiple languages (Python, C++, Java, etc.)~\cite{bostondynamics2020spotSDK} & Primarily C++ and Python within ROS \\ \hline
Ease of Integration & Direct access to robot functionalities via SDK & Seamless integration with ROS-based systems and tools \\ \hline
Community Support & Official support from Boston Dynamics & Community-driven support and development \\ \hline
Tools and Libraries & SDK-specific tools; requires custom development & Access to ROS tools (\texttt{rviz}, \texttt{rosbag}, \texttt{tf}) and libraries \\ \hline
\end{tabular}
\end{table}

For our system, ROS integration is advantageous due to its modularity and the ability to incorporate various perception and decision-making components within a unified framework. ROS provides middleware that simplifies communication between different system modules, including contextual HAR and behavior trees. While the Spot SDK with gRPC offers efficient, direct access to the robot's capabilities, integrating Spot with ROS allows us to leverage existing ROS packages and tools, facilitating development and interoperability with other systems.

\section{System architecture and workflow}

\begin{figure}[h]
    \centering
    \includegraphics[width=1.02\linewidth]{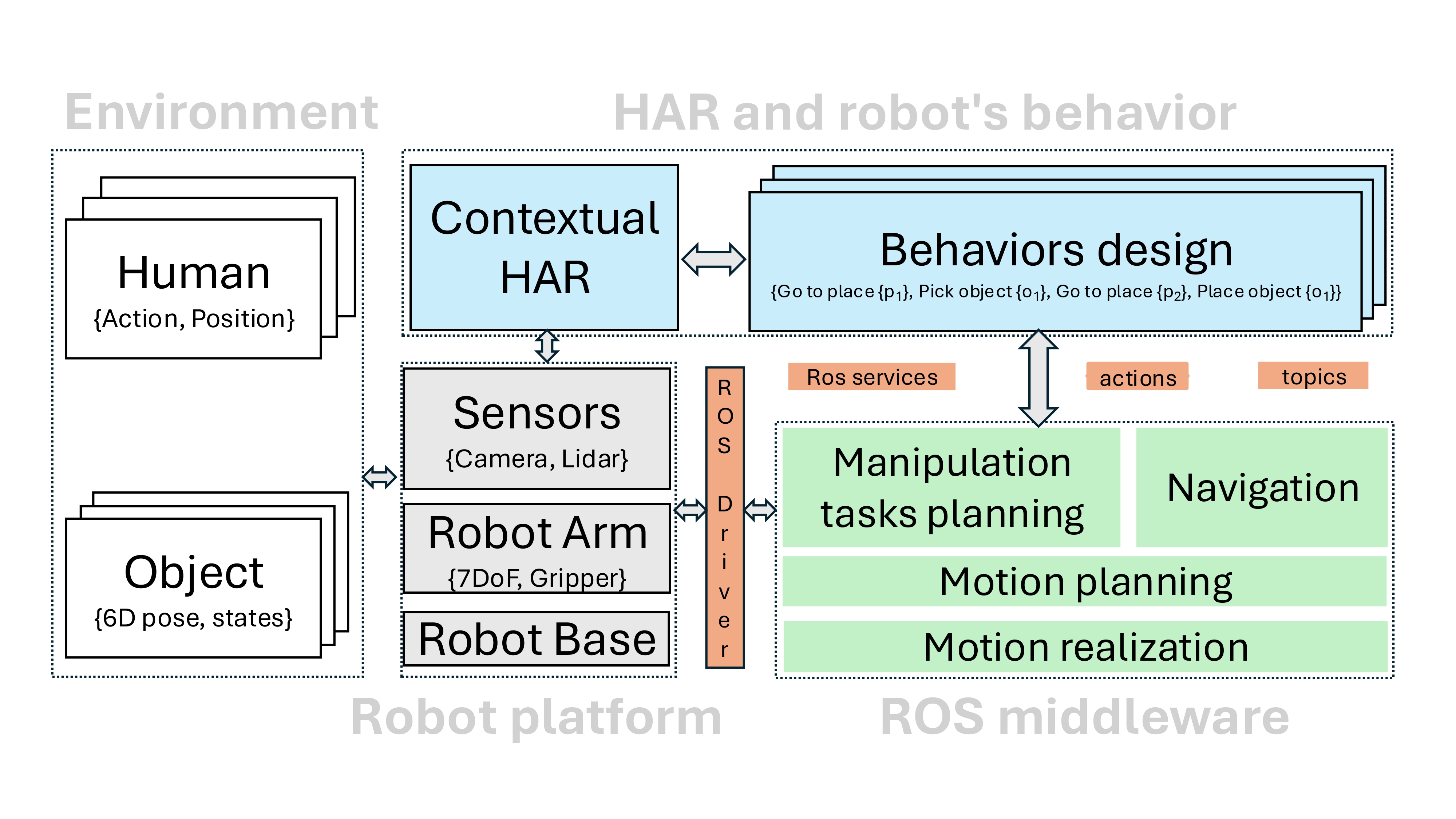}
    \caption{System's block diagram for proactive human-assistance robot using contextual HAR and behavior Trees}
    \label{fig:architecture}
\end{figure}

The proposed system integrates hardware and software components to create a context-aware, proactive human-assistance robot. The architecture includes the Boston Dynamics Spot robot platform, an edge computing device for real-time processing, and a software stack that combines the Robot Operating System (ROS), Behavior Trees (BTs), and Contextual Human Activity Recognition (HAR). An overview of this architecture is shown in Figure~\ref{fig:architecture}.

The architecture is organized into three main components: Environment, Robot Platform, and ROS Middleware, each essential for enabling responsive and adaptable interaction.

\subsection{Environment}

The environment includes both \textbf{humans} and \textbf{objects} that the robot interacts with. Human activities and positions are continuously monitored, providing essential data for proactive assistance. Objects in the environment are represented with six degrees of freedom (6D) poses and states to support precise manipulation and navigation. 

\subsection{Robot platform}

The robot platform integrates sensory and actuation components for perception, mobility, and environmental interaction. In our case, it leverages the Boston Dynamics Spot, a versatile quadruped robot equipped with advanced sensory and actuation components. Spot’s \textbf{sensor} suite includes high-resolution cameras and LiDAR, enabling precise environmental mapping, object detection, and human activity recognition. These sensors provide critical data for the contextual HAR module, supporting accurate recognition of human actions and tracking of object locations. Spot's 6 degrees of freedom (DoF) \textbf{robotic arm}, which includes an integrated gripper, allows Spot to manipulate various objects, performing tasks such as picking, placing, and handing items to humans, making it usable in assistive tasks. The \textbf{quadrupedal base} offers mobility and stability, enabling navigation across varied terrains and complex indoor settings. These components allow Spot to perceive, navigate, and interact within dynamic human environments. To integrate with ROS, Spot employs a dedicated \textbf{ROS driver} that standardizes control of its sensors, arm, and base within the ROS ecosystem. The driver accepts motion planner commands—specifying base motion, Cartesian pose of the end effector, and gripper opening width—and provides outputs from cameras and LiDAR to support contextual HAR, navigation, and manipulation planning. This driver enables efficient coordination between Spot’s hardware and the ROS-based software stack, as detailed in Section~\ref{sec:ros-mid}.

\subsection{ROS middleware} \label{sec:ros-mid}
\textit{The ROS middleware} is a module within our system architecture, acting as the bridge between high-level behavioral directives defined in BTs and their execution by the robot's hardware. This module transforms abstract task descriptions into precise, executable commands through several interconnected components, each specializing in distinct functionalities critical for the robot’s operations.

\paragraph{\textbf{Manipulation task planning}}
manages and coordinates the robot's interactions with objects—such as grasping, lifting, placing, and similar tasks—by leveraging advanced planning algorithms, including grasp synthesis algorithms \cite{1371616}. Using data-driven or analytical methods, these algorithms analyze object shapes and properties to identify stable and task-specific grasp points. By employing these approaches, the robot can execute complex tasks such as assembly, object retrieval, and precise object handling, even in dynamic and unstructured environments. \textbf{\textit{Navigation}} is enabled by the path planner, which calculates optimal routes to ensure collision-free operation within the environment. It combines heuristic and deterministic algorithms to dynamically compute paths, leveraging real-time environmental data from sensors such as cameras and LiDAR. Building on the output of the path planner, the \textbf{\textit{motion planner}} specifies detailed movements and configurations needed to achieve navigation and manipulation goals. This component uses kinematic and dynamic models to generate feasible and efficient motion trajectories, ensuring smooth transitions and adherence to the robot’s physical constraints. The \textbf{\textit{motion realization}} also known as the \textit{\textbf{motion control}} system translates motion plans into precise actuator commands. It uses advanced control algorithms to regulate motor outputs and incorporates feedback from proprioceptive sensors to adapt movements dynamically. This real-time feedback loop allows the robot to maintain stability, respond to unexpected disturbances, and operate effectively in dynamic environments.

Together, these components ensure the ROS middleware effectively translates high-level task plans into actionable commands by integrating navigation, manipulation, and control capabilities. The middleware also supports real-time data fed into the contextual HAR module, enhancing Spot’s understanding of human actions and environmental context. With this real-time feedback loop, Spot can dynamically adjust its responses to reflect ongoing human activities and environmental changes.

\subsection{System workflow and integration of contextual HAR and behavior trees}
The system workflow is designed as a cyclical, real-time process that combines contextual HAR and behavior trees to enable Spot’s proactive and adaptive response in dynamic environments.
The contextual HAR module continuously processes data from Spot’s sensory inputs such as cameras to interpret human actions and contextual information. This real-time perception enables Spot to recognize human activities that are used as inputs to the BTs.
Once the HAR module identifies a human action, the information flows to the BT module, which translates into actionable, high-level decisions. The BTs define structured sequences of actions, or subtrees, that allow Spot to respond proactively to human activities. For instance, upon detecting a \textit{reaching} action by a human, the BT can initiate a series of robot actions for Spot to move toward the user and prepare for \textit{object handover}. This hierarchical decision-making approach enables robots to execute tasks that are both dynamic and modular, allowing for rapid adaptation to a range of scenarios. After determining the appropriate actions through the BTs, the system utilizes the ROS middleware to convert these high-level decisions into specific commands for Spot’s base, arm, and gripper. The ROS middleware ensures that the actions are executed in coordination with the motion planner and manipulation modules, providing precise control over Spot’s movements and interactions. This integrated command execution allows Spot to interact with humans and objects.

\subsubsection{Contextual HAR:} \textit{contextual HAR} module is essential for enabling Spot’s proactive response to human actions. By processing data from Spot’s onboard cameras, the HAR module interprets human activities within their environmental context, allowing the robot to respond in alignment with human intent. The HAR process begins with the continuous acquisition of visual data, capturing human positions, movements, and interactions with objects. Key features—such as body orientation, gestures, and spatial positioning relative to objects—are extracted from this data to support activity recognition. The module leverages CNNs and RNNs to analyze spatial-temporal patterns, accurately identifying actions. A defining aspect of Contextual HAR is its capacity to integrate environmental context into action recognition. By assessing visual cues and object locations in the environment, the module can infer targets of human actions, such as when a reaching gesture is directed at a specific object. This context-driven approach ensures that Spot’s responses are appropriate and aligned with human intent.
\begin{figure}
    \centering
    \includegraphics[width=1\linewidth]{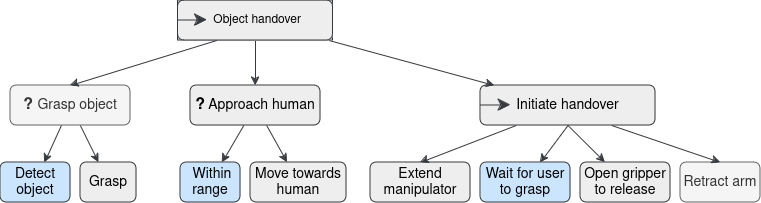}
    \caption{Behavior tree for object handover: Spot grasps the object, approaches the human, initiates the handover by extending the manipulator, waits for the user to grasp, and opens the gripper to release and retract the arm.}
    \label{fig:bt_detailed_handover}
\end{figure}

\subsubsection{Decision-making with behavior trees:}
\textit{BTs} form the core of decision-making, enabling robots to respond proactively to human activities from the contextual HAR module. Organized hierarchically, BTs consist of \textit{control nodes} and \textit{execution nodes}, each serving distinct functions. Control nodes, which include sequence nodes and selector nodes, direct the flow of actions within the tree. Sequence nodes ensure that a series of conditions or actions are performed in a strict order, advancing only after the previous node reports success. Selector nodes, on the other hand, provide alternative actions if the current one fails, thus ensuring that the system continually adapts to reach a viable solution.

Execution nodes, categorized into condition nodes and action nodes, perform specific tasks based on the robot's sensory data or internal state assessments. Condition nodes, depicted as blue blocks, evaluate whether the preconditions for subsequent action nodes, shown as gray blocks, are met. These action nodes are responsible for executing operational tasks such as moving, grasping, or navigating.

The integration of BTs with outputs from the HAR system allows for adaptive robot behavior based on the context of detected human activities. For instance, if the HAR system detects a person \textit{reaching} for a high shelf, the BTs adjust to prioritize supportive actions for retrieving items from that location, as demonstrated in Figure~\ref{fig:bt_detailed_handover}. The \textit{Grasp Object} selector node first ensures the object is grasped correctly, followed by the \textit{Approach Human} selector node that positions Spot optimally near the human. Once in place, the \textit{Initiate Handover} sequence node extends the manipulator and confirms the human's grip, allowing for a secure transfer of the object.

This decision-making architecture, detailed in BTs, empowers robotic systems like Spot to perform dynamically and adaptively in real time, responding appropriately to the complexities of human activity within varied environments.

\section{Conclusion}
This work presents a human-assistance robot system using Boston Dynamics Spot, enhanced by Contextual HAR and BTs within a ROS framework. By integrating HAR for real-time human action recognition with BTs for adaptive decision-making, the system enables Spot to respond proactively and contextually to human activities in dynamic environments. This framework supports complex tasks like object retrieval and handover, providing seamless, safe, and intuitive interaction. 
The use of BTs offers modularity and adaptability, allowing the robot to adjust its actions based on real-time HAR inputs. This flexibility is essential for effective assistance in unpredictable human environments, where conditions may change during task execution. The ROS middleware ensures efficient data flow between modules, maintaining system responsiveness and coordination.
This approach demonstrates the potential for deploying robots in assistive roles across settings like healthcare, workplaces, and personal environments. Future work could extend HAR capabilities with multi-modal data (e.g., auditory input) to capture a broader range of human activities and refine BTs with learning-based adaptations for more personalized assistance. Exploring deployment in diverse environments would further validate system robustness and scalability.
In summary, this research advances the field of human-assistive robotics by combining advanced perception and adaptive decision-making, paving the way for robots that seamlessly integrate into everyday life to enhance human safety, productivity, and quality of life.

%
%
\bibliographystyle{unsrt}
\bibliography{reference}
\end{document}